%% file: main.tex
\documentclass[10pt,twocolumn,letterpaper]{article}

\usepackage{cvpr}      % To produce the REVIEW version
\input{preamble}
\newif\ifarxiv
\arxivfalse

\definecolor{cvprblue}{rgb}{0.21,0.49,0.74}
\usepackage[pagebackref,breaklinks,colorlinks,allcolors=cvprblue]{hyperref}

\usepackage[scaled=.8]{beramono}
\usepackage[T1]{fontenc}
\usepackage{longfbox}
\usepackage{overpic}

\def\paperID{} % *** Enter the Paper ID here
\def\confName{CVPR}
\def\confYear{2026}

\title{\name: Language-Assisted Motion Planning for Controllable Video Generation}
\author{
Muhammed Burak Kizil\textsuperscript{1,*}
\and Enes Sanli\textsuperscript{1}
\and Niloy J. Mitra\textsuperscript{2,4}
\and Erkut Erdem\textsuperscript{3}
\and Aykut Erdem\textsuperscript{1}
\and Duygu Ceylan\textsuperscript{4}
\and {}
\\
{\small
\textsuperscript{1}Ko\c{c} University \quad
\textsuperscript{2}University College London \quad
\textsuperscript{3}Hacettepe University \quad
\textsuperscript{4}Adobe Research
}
}
\begin{document}

% \maketitle
\twocolumn[{%
\renewcommand\twocolumn[1][]{#1}%
\maketitle
\vspace*{-.3in}
\thispagestyle{empty}
% \vspace{-0.8cm}
\begin{center}
\centering 
\captionsetup{type=figure}
\includegraphics[width=1.0\linewidth]{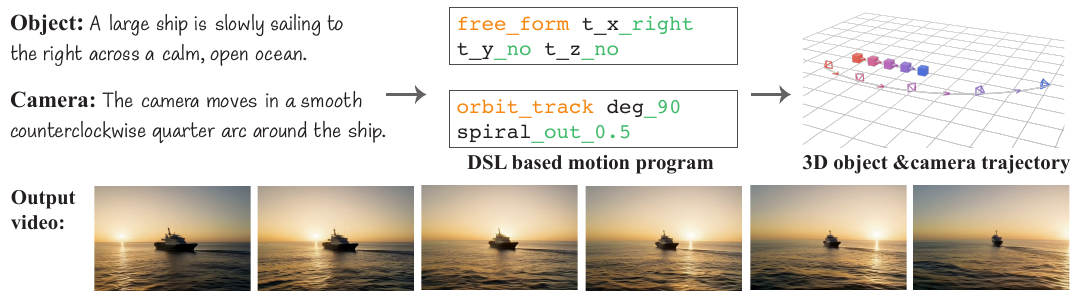}
%\vspace{1.5in}
\caption{\textbf{\name: from language to camera.} Given a natural-language description of a scene, \name interprets cinematic intent, how objects and cameras should move, and expresses it through a structured, cinematography-inspired language. The resulting symbolic motion programs are converted into 3D object and camera trajectories, bridging human language and machine-controllable motion. These trajectories condition a video generator to produce dynamic, visually coherent shots that faithfully reflect the described intent.}
%\captionof{figure}{\name introduces a cinematography inspired motion domain-specific language (DSL) and an LLM-based planner that synthesizes a symbolic motion program from a natural language description. These programs are deterministically converted into 3D object and camera trajectories which can be used for conditioning video generators.}%
\label{fig:teaser}%
\end{center}%
\vspace{1em}%
}]
\begingroup
  \renewcommand\thefootnote{}
  % Intern notu
  \footnotetext{$^*$This work was partially done while Burak was an intern at Adobe Research. }
\endgroup
\input{sections/00_abstract}
\input{sections/01_intro}
\input{sections/02_related}
\input{sections/03_method}
\input{sections/04_experiments}
\input{sections/05_conclusion}

\paragraph{Acknowledgements}
This work was partly supported by the KUIS AI Center Research Awards to M. Burak Kizil and Enes Sanli, and T\"{U}B\.{I}TAK-2247-A Program Award (No. 123C550) to A. Erdem. NM was partially supported by the UCL AI Centre.  We thank all the reviewers and Kenny Jones for their valuable comments.

\input{sections/06_supplemental}

\clearpage
\newpage
{
\small
\bibliographystyle{ieeenat_fullname}
\bibliography{main}
}

\end{document}

%% file: preamble.tex
\usepackage{amsmath}
\usepackage{amssymb}
\usepackage{booktabs}
\usepackage{dsfont}
\usepackage{graphicx}
\usepackage{makecell}
\usepackage{multicol}
\usepackage{multirow}
\usepackage{pifont}% http://ctan.org/pkg/pifont
\usepackage[group-separator={,}]{siunitx}
\usepackage{tabularx}
\usepackage{caption}
\usepackage{soul}
\usepackage{changepage,threeparttable}
\usepackage{colortbl}
\usepackage{algpseudocode}
\usepackage{algorithm}
\usepackage{xspace}
\usepackage{microtype}

\usepackage{array}
\usepackage[table,xcdraw]{xcolor}

\usepackage{mathtools}
\usepackage{xfrac}
\usepackage{url}

\usepackage{catchfile}
\usepackage{longtable}
\usepackage{tabu}
\usepackage{supertabular}
\usepackage{multicol}
\usepackage{enumitem}
\newcolumntype{Y}{>{\centering\arraybackslash}X}
\usepackage[accsupp]{axessibility}  % Improves PDF readability for those with disabilities.

\definecolor{color_1}{RGB}{255,0,128}
\definecolor{color_2}{RGB}{0,128,128}
\definecolor{color_3}{RGB}{0,128,0}
\definecolor{color_4}{RGB}{128,0,0}
\definecolor{color_5}{RGB}{128,0,128}
\definecolor{cadetgrey}{RGB}{0.57, 0.64, 0.69}

\newcommand{\name}{LAMP\xspace}
\newcommand{\motion}{primitive\xspace}
\newcommand{\users}{16\xspace}

\renewcommand{\paragraph}[1]{{\vspace{1.5mm}\noindent \bf #1}.}

%% file: sections/00_abstract.tex
\begin{abstract}

Video generation has achieved remarkable progress in visual fidelity and controllability, enabling conditioning on text, layout, or motion. Among these, motion control -- specifying object dynamics and camera trajectories -- is essential for composing complex, cinematic scenes, yet existing interfaces remain limited. %Current approaches struggle to capture the inherently entangled relationship between object motion and camera behavior. 
We introduce \name that leverages large language models~(LLMs) as motion planners to translate natural language descriptions into explicit 3D trajectories for dynamic objects and (relatively defined) cameras. %Specifically, we fine-tune an LLM to generate framewise 3D bounding-box trajectories for objects and, conditioned on these, produce corresponding 3D camera paths, 
%These trajectories are converted into generator-compatible 2D control signals. 
\name defines a motion domain-specific language (DSL), inspired by cinematography conventions. By harnessing  program synthesis capabilities of LLMs, \name generates structured motion programs from natural language, which are deterministically mapped to 3D trajectories. We construct a large-scale procedural dataset pairing natural text descriptions with corresponding motion programs and 3D trajectories. %text–trajectory pairs and augmented real video datasets with 3D annotations. %By decoupling trajectory planning from synthesis, our framework enables efficient user iteration via relative text prompts (e.g., “zoom out more”) or direct 3D manipulations before costly generation. 
Experiments demonstrate \name's improved performance in motion controllability and alignment with user intent compared to state-of-the-art alternatives~\cite{courant2024et,zhang2025gendop}, establishing the first framework for generating both object and camera motions directly from natural language specifications. 
Video and code are available at the project page at \url{https://cyberiada.github.io/LAMP/}.

\end{abstract}

%% file: sections/01_intro.tex
\section{Introduction}
%Video generation and the importance of dynamics
Video generation has emerged as a powerful medium for creative expression and visual storytelling. Foundational video generators~\cite{kong2024hunyuanvideo,yang2024cogvideox,wan2025} have advanced rapidly, not only in visual fidelity but also in providing multimodal controls. Beyond text- and image-guided synthesis, recent models can now incorporate conditioning on structures (e.g., depth), camera parameters~\cite{cameractrl2,epic}, and motion signals~\cite{vace}, offering unprecedented opportunities to direct complex scenes. Among these, \textit{motion control}, i.e., governing both object trajectories and camera paths, is especially critical, as it provides the directorial leverage needed to shape how stories unfold and to generate cinematic, physically consistent, and narratively expressive videos

%Limitations of current controls
Despite the expanded control capabilities of modern video generators, aligning generated outputs with user intent remains difficult. We argue that the bottleneck lies not in the models themselves, but in the limited ways users can specify control signals. Today, most interaction still relies on text, annotations extracted from existing videos~\cite{yang2023effective,video_depth_anything}, or simple 2D sketching interfaces~\cite{geng2024motionprompting}. These modalities are restrictive, especially when describing choreographed motion and complex directorial camera. In many cases, \textit{object dynamics and camera trajectories are inherently entangled}, i.e., camera is usually defined relative to the moving objects. Simultaneously specifying \textit{both} requires advanced spatial planning and mental imagination, far beyond the reach of casual users. For example, choreographing a chase scene demands careful coordination of both the runner’s path and the pursuing camera. Specifying all this in terms of depth or motion signals is difficult to mentally imagine and coordinate in time.

We address this gap by introducing \name (Language-Assisted Motion Planning), a framework that treats motion control as a language-to-program synthesis problem. Instead of regressing raw coordinates, LAMP leverages the reasoning abilities of large language models (LLMs) to generate symbolic motion programs in a domain-specific language (DSL) inspired by cinematography. The DSL offers an interpretable and compositional interface that represents both object and camera behavior through motion primitives (e.g., orbit, tail-track, rotation-track) and modifiers (e.g., speed, easing, offset). These programs are deterministically mapped to 3D trajectories, ensuring spatial consistency and enabling iterative refinement. Unlike prior approaches that focus solely on layout generation~\cite{Lin2023VideoDirectorGPT} or camera trajectory synthesis~\cite{zhang2025gendop}, \name unifies object and camera motion planning in a shared 3D space, a key requirement for cinematographically coherent video synthesis (see \cref{fig:teaser}).

%
%Framework overview
%Concretely, we fine-tune an LLM~\cite{qwen2.5-VL} to generate 3D object trajectories represented as frame-wise bounding boxes with positions and orientations. Conditioned on the textual description and the generated object trajectories, the LLM then produces a corresponding 3D camera path. These trajectories are subsequently converted into generator-compatible formats (e.g., 2D bounding boxes) along with moving horizon lines, enabling controllable synthesis. For example, in \Cref{fig:teaser} we show ... 

%Dataset construction
%Concretely, we finetune an LLM~\cite{qwen2.5-VL} to generate symbolic motion programs from natural language both for moving objects and the camera. To support this finetuning, we construct a procedural dataset that pairs natural language descriptions with corresponding motion programs and 3D trajectories, which we will make publicly available. %We procedurally generate large sets of trajectory–text pairs using an LLM, and further combine this with real-world video datasets~\cite{courant2024et}. This combination provides the necessary supervision for joint trajectory generation.
% 
%\Niloy{mention that we will release the dataset?}
%
To train the LLM motion planner, we construct a large-scale procedural dataset comprising 400K paired text–motion samples, where each natural-language description is aligned with a symbolic DSL program and its corresponding 3D trajectories. The dataset is generated automatically by sampling and composing motion primitives, and an auxiliary LLM is used to paraphrase each description for linguistic diversity. This approach scales to diverse, physically coherent motion patterns and enables generalization to free-form text prompts. We then fine-tune a vision–language LLM~\cite{qwen2.5-VL} on this corpus to produce symbolic motion programs directly from natural-language inputs, covering both object and camera behaviors. The resulting dataset and models will be publicly released to support future research on language-driven motion planning.

By decoupling motion planning from the underlying video generator, \name allows efficient and editable control, users can issue relative refinements such as “move the camera lower” or “zoom out slightly” \textit{before} committing to costly (video) synthesis.
%Evaluation
To the best of our knowledge, ours is the first framework to generate both object and camera trajectories from natural language specifications. We evaluate \name on existing benchmarks~\cite{courant2024et}, and demonstrate consistent improvements in motion controllability, text-trajectory alignment, and perceptual plausibility. Qualitatively, LAMP produces coherent and cinematic camera-object coordination, while quantitatively it mostly achieves superior results compared to state-of-the-art baselines.  
 
In summary, our main contributions are:
(i)~\textit{A cinematography-inspired motion domain-specific language (DSL)} that encodes camera and object behaviors as symbolic motion programs deterministically mapped to 3D trajectories; 
%(ii)~\textit{A large-scale procedural text–motion dataset}, enabling scalable training and benchmarking of language-driven motion planners; 
(ii)~\textit{An open procedural data generation framework and large-scale text–motion dataset}, enabling scalable creation, training, and benchmarking of language-driven motion planners 
(iii)~\textit{flexible interface:} a decoupled motion planning stage that supports iterative refinement via text before actual video synthesis.
Extensive experiments demonstrate that \name provides  superior camera and layout controllability compared to recent baselines~\cite{courant2024et,zhang2025gendop} while also enabling iterative motion refinement before actual video synthesis.  

%% file: sections/02_related.tex
\vspace{-2mm}
\section{Related Work}
\vspace{-2mm}
\paragraph{Video generation and control}
Recent text-to-video (T2V) diffusion models have achieved impressive progress in visual fidelity, temporal coherence, and prompt alignment~\cite{wan2025, agarwal2025cosmos, veo2_2025, openai2024sora}. However, these models lack explicit understanding of motion, often producing plausible, yet physically inconsistent dynamics. To address this, several approaches introduced structural and motion conditioning: VidCRAFT3~\cite{vidcraft3} jointly models lighting and motion via a spatial triple-attention transformer; FloVD~\cite{flovd} leverages optical flow to constrain background and camera dynamics; while I2VControl~\cite{i2vcontrol} unifies camera, depth, and trajectory cues for multi-modal control. RoPECraft~\cite{gokmen2025ropecraft} enables training-free motion transfer by injecting optical-flow-guided updates into rotary positional embeddings (RoPE) of diffusion transformers. These methods, however, treat motion as a passive conditioning cue, limiting coordination between objects and cameras. \name instead casts motion control as a \emph{language-to-trajectory planning} task, using an LLM to govern the movement of both the object and camera.

\paragraph{Cinematography and camera control}
Camera motion has become a central focus in controllable video generation. CineMaster~\cite{cinemaster} and RealCam-I2V~\cite{realcam} introduced 3D-aware pipelines that enable direct manipulation of virtual cameras;  CameraCtrl~II~\cite{cameractrl2} and ReCamMaster~\cite{recammaster} extended these ideas to dynamic and multi-camera settings; LightMotion~\cite{lightmotion} and TrajectoryCrafter~\cite{trajectorycrafter} propose lightweight conditioning strategies for motion-guided generation; SAGE~\cite{kan2026sage} for cinematographic inbetweening;  EPiC~\cite{epic} improves efficiency and precision through visibility-based anchor-video construction, removing the need for camera trajectory annotations. GEN3C~\cite{gen3c} and Stable Virtual Camera~\cite{seva} integrate explicit or implicit intermediate 3D feature representations that maintain world consistency and enable long-horizon synthesis; while CamCloneMaster~\cite{camclonemaster} enables reference-based camera cloning without pose annotations. Most such approaches, however, emphasize camera-side control while assuming static or implicitly modeled objects. As a result, they do not coordinate moving subjects and cameras jointly. Ours bridges this gap by generating both \emph{object--camera trajectories} in a shared 3D space, and hence cinematographically grounding motion planning. Furthermore, the camera motion paths generated by \name can potentially be used as input to many of these prior approaches.

\paragraph{LLMs as planners}
As video generators become increasingly capable of conditioning on motion cues, a parallel line of work explores how to derive such cues directly from text, with most efforts focusing on camera motion. Diffusion-based methods for predicting 3D camera trajectories~\cite{CCD,Director3D,courant2024et} produce plausible results, while recent advances show that LLMs possess strong reasoning and planning capabilities across robotics, navigation, and embodied AI. Their use in video generation, however, remains limited. Modular-Cam~\cite{modularcam} employs an LLM (the ``LLM-Director'') to parse complex instructions into scene descriptions and transition actions, selecting among a small library of pretrained CamOperator modules (e.g., zoom, pan) to assemble a camera path. VideoDirectorGPT~\cite{Lin2023VideoDirectorGPT} similarly uses an LLM to expand a prompt into a multi-scene video plan with entities, layouts, and consistency groups, which Layout2Vid then grounds visually, yet the LLM never outputs explicit 3D geometric trajectories. Recent work begins to close this gap. ET~\cite{courant2024et} introduces a large-scale set of paired camera–character trajectories and captions, using LLM-generated cinematic descriptions to guide a diffusion-based trajectory model. GenDoP~\cite{zhang2025gendop} uses GPT-based models to produce detailed directorial and motion captions for autoregressive camera path generation. These approaches show that LLMs can encode high-level cinematographic semantics, though their role is still largely restricted to captioning or auxiliary supervision rather than direct motion planning. 
Instead, we fine-tune an LLM to directly output motion programs for both objects and cameras. By coupling linguistic intent with explicit geometric control, \name offers a scalable and editable interface for specifying complex, semantically grounded cinematic motion.

%% file: sections/03_method.tex
\section{Method}
\subsection{Overview}
\begin{figure}[t!]
    \centering
    \includegraphics[width=\columnwidth]{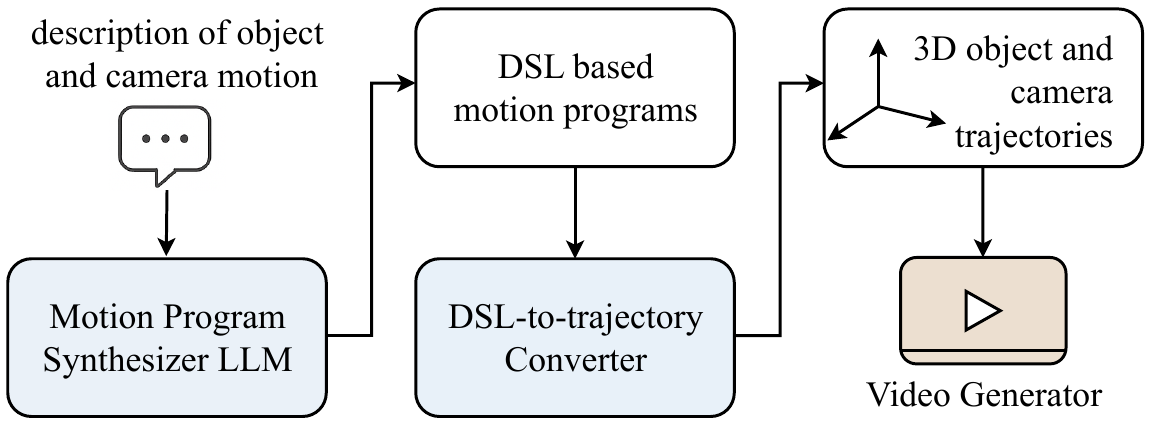}
    \caption{
    \textbf{Overview of \name.} A learned LLM acts as a motion planner, generating symbolic motion programs in a DSL format from textual descriptions of object and camera motion. These programs are deterministically converted into 3D trajectories, which are used to condition a pretrained video generator.
    }
    \label{fig:overview}
\end{figure}

Our objective is to translate natural-language descriptions into explicit, geometrically consistent trajectories for \textit{both} dynamic objects and the associated camera. We propose \name (Language-Assisted Motion Planning), a framework that employs an LLM as a motion planner to generate explicit, cinematographically meaningful object and camera trajectories from text. Given a natural-language description $\mathbf{t}$, the motion planner $\mathcal{F}_\theta$, a finetuned autoregressive LLM, produces symbolic motion programs for both the object and the camera, denoted as $(\mathbf{s}_{\text{obj}}, \mathbf{s}_{\text{cam}})$. These programs are expressed in a structured Domain-Specific Language (DSL) derived from cinematographic conventions and are deterministically mapped to 3D object and camera trajectories $(\mathbf{x}_{\text{obj}}, \mathbf{x}_{\text{cam}})$ within a shared world coordinate system. The resulting trajectories are rendered as a control video, which is then used to condition a pretrained video diffusion model $\mathcal{G}$ to synthesize the final video (Fig. \ref{fig:overview}).

\name unites three key ideas: (i) a language-native motion planner that interprets spatial and temporal semantics from text, (ii) a cinematography-grounded DSL extending the CameraBench~\cite{camerabench} taxonomy to encode canonical camera motion primitives and compositional modifiers, and (iii) a deterministic mapping from symbolic motion to 3D geometric hints for controllable video generation.
To represent scene motion compactly, we use 3D bounding boxes whose trajectories encode position over time. This abstraction mirrors standard previsualization and blocking practices in filmmaking, where simplified geometric proxies are employed to reason about staging, shot composition, and camera movement prior to rendering.

%We present \name, the first framework for object and camera trajectory generation from natural language descriptions. \name is a finetuned autoregressive generator, $\mathcal{F}$, that is finetuned from a large language model (LLM). $\mathcal{F}$ takes as input a text description $\mathcal{T}$ and generates the corresponding camera $\mathcal{C}_s$ and object $\mathcal{O}_s$ motion sequences (Section~\ref{sec:llm}). Inspired by the success of LLMs in program synthesis~\cite{Ma2025mover}, we define a \emph{domain specific language} (DSL) to represent these motion sequences (Section~\ref{sec:dsl}). Given the camera and object motion sequences generated by the LLM, we next use a deterministic \emph{DSL-to-trajectory} function to obtain the corresponding object and camera trajectories $\mathcal{O}_t$ and $\mathcal{C}_t$ in a common world coordinate system (Section~\ref{sec:traj}). Each such trajectory is generated by converting the LLM-generated motion language to per-frame 3D position and orientation information. Finally, we use the 3D trajectories $\mathcal{O}_t$ and $\mathcal{C}_t$ to render a control video which we then provide as conditioning signal to a video generator $\mathcal{G}$. Given a text prompt, the control video, and an optional first frame image, $\mathcal{G}$ generates the final video.

\begin{figure*}[!t]
    \centering
    \includegraphics[width=0.92\textwidth]{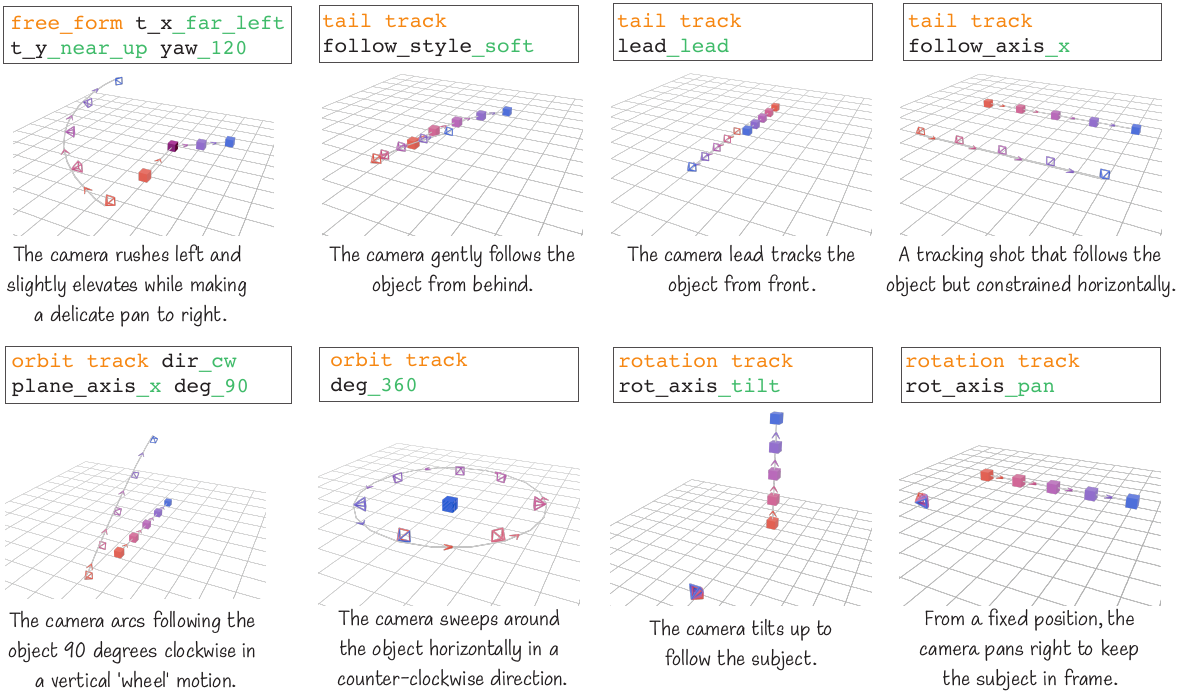}
    \caption{
    \textbf{Motion primitives.} Examples of the primitive camera motion types encoded by our DSL. Each illustration depicts a distinct primitive with a natural-language description of the corresponding cinematic motion. The DSL (shown in boxes) specifies the \textcolor{orange}{primitive type} followed by a list of modifiers in key\textcolor{teal}{\_value} format. See supplemental for the full DSL specification and many more examples. 
    }
    \label{fig:DSL}
\end{figure*}

\subsection{Motivation and Design Rationale}
Directly regressing 3D trajectories for cameras and objects from text is inherently difficult: the mapping from language to motion is multi-modal, structurally constrained, and suffers from a lack of large paired data. Rather than predicting continuous coordinates, we instead prompt the LLM to \textit{generate symbolic motion programs} composed of interpretable commands. This design is inspired by the language of cinematography, where movement is expressed through a compact, composable vocabulary such as pan, \textit{tilt}, \textit{truck}, \textit{orbit}, or \textit{zoom} that naturally lends itself to formalization as a language.
Building on the CameraBench taxonomy~\cite{camerabench}, which organizes canonical camera motion primitives across camera-, ground-, and object-centric reference frames, we design a motion domain-specific language (DSL) that encodes these primitives and their modifiers. This structured representation offers several advantages:
\begin{enumerate}[i)]
\item \textbf{Data efficiency:} Symbolic composition replaces frame-level supervision, reducing the need for large annotated datasets.
\item \textbf{Interpretability:} Each DSL command corresponds to a concrete cinematic operation, enabling inspection and editing.
\item \textbf{Compositionality:} Complex motions emerge from short, reusable programs built from simple primitives.
\end{enumerate}
Consequently, \name reframes motion generation as language-conditioned program synthesis, where the LLM learns to articulate cinematic intent using the same motion grammar employed by human directors. This abstraction mirrors standard previsualization and blocking practices in cinematography~\cite{Bordwell2020,Katz1991} and aligns with established geometric camera planning approaches in computer graphics~\cite{Christie2008,Lino2015}.

\newfboxstyle{rtight}{padding=1pt,margin=0.5pt,baseline-skip=false}

\subsection{Motion Domain Specific Language}
\label{sec:dsl}
\paragraph{Syntax} Each motion program is represented as an ordered list of motion tags: 
\lfbox[rtight]{\texttt{\motion}} \lfbox[rtight]{\texttt{modifier\_1}} \lfbox[rtight]{\texttt{modifier\_2}} \dots \lfbox[rtight]{\texttt{modifier\_n}},
where the \motion specifies the base motion primitive and the subsequent modifiers refine its parameters. Any unspecified modifier defaults to its canonical~value.

\paragraph{Base Primitives} Inspired by CameraBench’s primitive taxonomy, our DSL defines four canonical motion primitives that together cover most cinematic operations such as following, revealing, and encircling a subject (see Fig.~\ref{fig:DSL}):
\begin{enumerate}
\item \textbf{Free-form:} Unconstrained 6-DoF motion for either objects or cameras.
\item \textbf{Orbit track:} The camera revolves around a target object at a fixed or variable radius.
\item \textbf{Tail track:} The camera follows an object with a configurable temporal offset.
\item \textbf{Rotation track:} The camera rotates in place to maintain framing or reveal the scene.  
\end{enumerate}

\vspace{-1mm}
\paragraph{Modifiers} Each motion primitive is parameterized by translational controls \texttt{(lat, vert, depth)}, rotational controls \texttt{(yaw, pitch, roll)}, and temporal or stylistic cues such as \texttt{(speed\_fast, ease\_in, jitter\_low)}. Modifiers are expressed as key–value pairs \texttt{(k, v)}, forming a concise grammar that is both human-readable and machine-parsable. A complete list of supported modifiers for each primitive is provided in the supplementary.

\vspace{-1mm}
\paragraph{Sequence Representation} An object motion sequence $\mathbf{s}_{\text{obj}} := \{s^0_{\text{obj}},\dots,s^{N-1}_{\text{obj}}\}$ consists of up to $N = 4$ motion tags, limited to translation and  pitch and yaw for simplicity. Camera motion sequences $\mathbf{s}_{\text{cam}} := \{s^0_{\text{cam}},\dots,s^{N-1}_{\text{cam}}\}$ allow full 6-DoF control. Similar to the objects, we use up to four tags for free-form motion.

\subsection{Procedural Text-Trajectory Corpus}
%To train the motion planner $\mathcal{F}$, we construct a procedural corpus of text–motion examples covering a broad range of object and camera behaviors. Each motion sequence spans $T=21$ frames and is divided into four temporal segments, where we randomly sample whether motion primitives change across segments: 35\% single-primitive, 30\% two-primitive, and 35\% multi-primitive sequences.

%Each motion primitive is associated with multiple textual templates (e.g., \textit{``the camera pans slowly upward''}, \textit{``a handheld jitter follows the subject''}, \textit{``the drone orbits clockwise around the object''}). To enrich linguistic diversity, we employ an auxiliary LLM to paraphrase and extend each caption, producing natural, varied textual descriptions that mirror real cinematographic instructions.

%Finally, we employ a deterministic DSL-to-trajectory converter to map symbolic motion sequences into 3D trajectories. For free-form motion, translation and rotation parameters are integrated per frame relative to the origin. For tracking behaviors, the object trajectory is generated first, and the camera pose is then computed relative to the object’s position and orientation before being transformed into world-space coordinates. This process produces physically coherent world-aligned trajectories $(\mathbf{s}_{\text{obj}}, \mathbf{s}_{\text{cam}})$ paired with linguistically rich textual annotations. The resulting corpus provides scalable supervision for fine-tuning the LLM motion planner, ensuring broad coverage across diverse cinematic scenarios.

To train the motion planner $\mathcal{F}$, we construct a large-scale procedural corpus of text–motion examples covering a broad range of object and camera behaviors. Each motion sequence spans $T=21$ frames and is divided into four temporal segments, where motion primitives may change across segments. For both free-form and object-relative motion, we generate symbolic DSL programs paired with textual descriptions. We construct a procedural corpus of 400K text–motion pairs, comprising 100K samples each for free-form and object-relative motions (both raw and LLM-paraphrased). Text templates describe canonical cinematographic behaviors (e.g., ``the camera pans slowly right’’, ``the drone orbits clockwise around the object’’), and an auxiliary LLM paraphrases each template to further expand linguistic diversity.

Given a symbolic DSL program, we apply a deterministic DSL-to-trajectory converter to obtain physically coherent 3D motion. For free-form primitives, translation and rotation parameters are integrated per frame relative to the origin. For tracking primitives, the object trajectory is synthesized first, and the camera trajectory is computed relative to the object before being transformed into world coordinates. This yields world-aligned object–camera trajectories $\mathbf{s}_\text{obj}$ and $\mathbf{s}_
\text{cam}$,paired with natural-language descriptions.

To characterize the diversity of the free-form motions, we analyze both coarse and fine-grained motion-tag distributions for objects and cameras. The corpus spans 27 coarse classes (3 motion types × 3 directions) and 343 fine-grained classes (7 motion types × 3 directions) for translation. Moreover, rotations for camera are nearly unconstrained, sampled densely across the full $[-180, 180]$ degree range. This serves as a primary source of motion diversity, as the resulting orientation also influences the camera's translational path. As detailed in the supplementary material, the distribution is controllable and intentionally imbalanced, with common cinematographic motions (e.g., forward, backward) appearing frequently and multi-axis composite motions appearing less often. This pattern mirrors real-world datasets such as E.T.~\cite{courant2024et} and GenDoP~\cite{zhang2025gendop}. The resulting coverage ensures that the LLM planner encounters both frequent primitives and diverse long-tail combinations, enabling generalization to complex, multi-primitive cinematic behaviors.

\subsection{LLM-Based Motion Planning}
\label{sec:llm}
At inference time, the motion planner $\mathcal{F}$ receives a natural-language prompt $\mathbf{t}$ and produces symbolic motion programs for both the object and the camera in DSL form.
We first decompose the input description  $\mathbf{t}$ into object- and camera-centric components, $\mathbf{t}_\text{obj}$ and $\mathbf{t}_
\text{cam}$, and model the joint probability distribution as:  
\begin{equation}
p(\mathbf{s}_{\text{cam}}, \mathbf{s}_{\text{obj}} | \mathbf{t})
:= p(\mathbf{s}_{\text{obj}} | \mathbf{t}_{\text{obj}})\,
p(\mathbf{s}_{\text{cam}} | \mathbf{s}_{\text{obj}}, \mathbf{t}_{\text{cam}}),
\end{equation}
\noindent anchoring the coordinate system to the first camera frame. This factorization reflects the hierarchical structure of cinematography: the subject's motion defines the scene dynamics, while the camera, conditioned on the subject, adjusts to preserve framing and narrative continuity.

\begin{figure}[!h]
    \centering
    \includegraphics[width=\columnwidth]{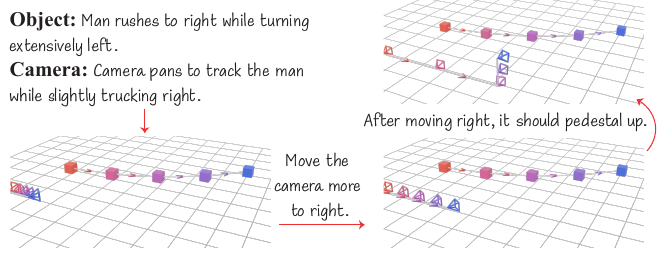}
    \caption{\name allows iterative control over synthesized motion trajectories via simple textual instructions. We visualize the 3D trajectories converted from the LLM predicted motion programs.}
    \label{fig:iterative}
\end{figure}

The LLM generates motion tags, conditioned on previously generated tokens and context. This symbolic formulation enforces long-range temporal consistency, produces syntactically valid DSL programs, and allows for explicit human-in-the-loop refinement, where users can iteratively modify motion descriptions as shown in Fig.~\ref{fig:iterative}.

%\subsection{LLM Guided Trajectory Generation}
%\label{sec:llm}
%Given a text description of object and camera motions, \name predicts the corresponding motion sequences in the domain specific language format described before. We assume that the text description $\mathcal{T}$ provided by the user consists of two components, namely $\mathcal{T}_o$ and $\mathcal{T}_c$, that describe the object and camera motions respectively. 

%How to capture the distribution of camera with object with movement? we model the scene in the world coordinate (set as the first frame of the camera); then, we have a conditional distribution of camera given object. All this, given text prompts.
%
%
%Modeling the joint distribution
%
%
\subsection{Trajectory-Conditioned Video Generation}
\label{sec:traj}
We convert the LLM predicted motion programs in the DSL form to 3D trajectories using the deterministic DSL-to-trajectory converter similar to our procedural dataset construction. We render the predicted trajectories into a dense, frame-aligned control video compatible with diffusion-based video generators such as VACE \cite{vace}.
For each frame, 3D object bounding boxes are projected into the current view as 2D overlays, while the edges of a fixed global cube are projected to represent horizon and camera orientation. %The resulting composited frames encode both object and camera dynamics as a dense, frame-aligned control signal. 
Given the text prompt, control video, and optionally the first frame, the pretrained generator $\mathcal{G}$ synthesizes the final video that follows the intended cinematic motion. %\hl{During training, we apply small spatial shifts and scale perturbations to improve robustness across varying scene layouts and viewing configurations.}

%% file: sections/04_experiments.tex
\section{Evaluation}
We evaluate \name across a wide range of quantitative benchmarks and qualitative visual results to assess its effectiveness in translating language into coordinated object–camera motion. Our experiments focus on three main aspects: (i)~accuracy and realism of camera trajectory generation; (ii)~fidelity of object motion prediction; and (iii)~perceptual and cinematic quality of the final synthesized videos. We also include ablation studies to isolate the effect of the proposed DSL and fine-tuning strategy, and user studies to evaluate perceived realism and alignment with textual intent. We refer to the supplementary material for extensive visual and quantitative results.

\subsection{Camera Motion Evaluation}
%We evaluate the ability of \name to translate textual descriptions into plausible and coherent camera motion sequences, and compare its performance against recent text-guided camera trajectory generation methods.

\paragraph{Baselines} We compare \name with representative text-conditioned camera trajectory models, including the diffusion-based CCD~\cite{CCD}, ET~\cite{courant2024et}, and Director3D~\cite{Director3D}, as well as the most recent autoregressive approach, GenDoP~\cite{zhang2025gendop}. While these baselines directly regress per-frame camera coordinates (discretized into bins in the case of GenDoP), our method first predicts a structured motion program in DSL form, which is then deterministically converted into explicit 3D camera trajectories. This symbolic formulation ensures spatial consistency and interpretability, while preserving the precision of coordinate-level control.

\paragraph{Metrics}
For evaluation, we adopt the Contrastive Language–Trajectory (CLaTr) embedding proposed in ET~\cite{courant2024et}, which measures how well generated trajectories align with textual descriptions. Specifically, we use CLaTr-CLIP to quantify text–trajectory alignment in a shared embedding space and CLaTr-FID to assess the overall realism and naturalness of the generated motions. In addition, we report the motion tagging F1 score, proposed in~\cite{courant2024et}, to evaluate the accuracy with which the predicted trajectories capture the semantic camera actions described in text.

\paragraph{Results} We conduct quantitative evaluations across multiple datasets to assess \name for generating text-aligned camera motions. Table~\ref{tab:datadop} presents results on the DataDoP dataset introduced by GenDoP~\cite{zhang2025gendop}. Since the official train/test split has not been released, we evaluate \name on a representative subset of publicly available data. For fairness, we also evaluate the released GenDoP checkpoint on the same subset, denoted as \emph{GenDoP (exp)}. Although trained without access to the original DataDoP training set, \name (denoted as \emph{Ours (DSL)}) achieves superior performance compared to models explicitly trained on this dataset, such as GenDoP and ET, and significantly outperforms all pretrained baselines. We also convert a filtered subset from DataDoP to DSL format for additional finetuning (last row) which further improves the performance. This demonstrates that our structured DSL representation generalizes well to unseen motion domains. %We refer to the supplementary for more details and results.
\begin{table}[!h]
\centering
\caption{\textbf{Camera trajectory evaluation on the DataDoP dataset.} \name achieves superior performance than DataDoP-trained baselines despite no dataset-specific training.}
\label{tab:datadop}
\resizebox{\linewidth}{!}{
\begin{tabular}{rc@{$\;\;$}c@{$\;\;$}c@{$\;\;$}c@{$\;\;$}c@{$\;\;$}c}
\toprule
\textbf{Model} & \textbf{Data} & \textbf{Revised F1-Score}&\textbf{F1-Score}  & \textbf{CLaTr} & \textbf{Coverage} & \textbf{FID} \\
\midrule 
CCD & pretrained &- &0.297 & 5.29 & 0.332 & 357.822 \\
ET & pretrained & -&0.330 & 2.46 & 0.020 & 609.906 \\
Director3D & pretrained& -& 0.058 & 0.00 & 0.171 & 542.385 \\
Director3D & DataDoP & -&0.391 & 31.69 & 0.839 & 31.979 \\
GenDoP  & DataDoP &- &\textbf{0.400} & \textbf{36.18} & \textbf{0.872} & \textbf{22.714} \vspace{0.15cm}\\
GenDoP (exp) & DataDoP & 0.360 &- & 35.91 &\textbf{0.853} & \textbf{48.123}\\
Ours & pretrained & \textul{0.763} &- & \textul{36.29} & \textul{0.794} & \textul{66.86} \\
Ours & (ft) w/ DataDoP & \textbf{0.776} & - & \textbf{36.52} & 0.779 & 67.24 \\
\bottomrule
\end{tabular}
}
\end{table}

Table~\ref{tab:et} reports results on the ET dataset~\cite{courant2024et}, which provides two evaluation splits: a simpler (\emph{pure}) set and a more complex (\emph{mixed}) one. \name consistently outperforms all baselines across both splits, despite being trained on a different dataset. We also convert a filtered subset from ET to DSL format for additional finetuning (last row) which further improves the performance. %Our model maintains high motion fidelity in the mixed setting, confirming that the proposed LLM-based planner effectively transfers across motion domains and supports complex cinematographic behaviors.
\begin{table}[!t]
\centering
\caption{\textbf{Camera trajectory results on the ET dataset.} The ET benchmark includes a simpler \emph{pure} and a harder \emph{mixed} split. 
\name attains the highest F1-scores on both, demonstrating strong generalization across (unseen) motion complexity levels.}
\label{tab:et}
\resizebox{0.95\linewidth}{!}{

\begin{tabular}{r c@{$\;\;$}c@{$\;\;$}c@{$\;\;$}c}
\toprule

\multirow{2}{*}{\textbf{Model}} & \multicolumn{2}{c@{$\;\;$}}{\textbf{Pure Split}} & \multicolumn{2}{c}{\textbf{Mixed Split}} \\
\cmidrule(lr){2-3} \cmidrule(lr){4-5} % Pure ve Mixed'in altına çizgi ekler
& \textbf{F1 Score} & \textbf{CLaTr Score} & \textbf{F1 Score} & \textbf{CLaTr Score} \\

\midrule 

CCD & 0.27 & 3.21 & 0.17 & 6.26 \\
MDM & 0.76 & 21.26 & 0.34 & 18.32 \\
ET - DirB & 0.86 & 23.10 & 0.39 & 20.78 \\
ET - DirC & 0.80 & 21.49 & 0.48 & 21.95 \\
Ours & \textul{0.976} & \textbf{35.10} & \textul{0.769} & \textul{36.59} \\
Ours (ft w/ ET) & \textbf{0.978} & \textul{35.02} & \textbf{0.779} & \textbf{36.95} \\
\bottomrule
\end{tabular}
}
\end{table}

\subsection{Object Trajectory Evaluation}
 We evaluate the ability of \name to generate text-aligned object trajectories using motion tagging on the test split of our procedural dataset. For translation, we consider both 27 coarse motion classes and 343 fine motion classes. For rotation, we evaluate across 25 motion classes, as for the objects rotation options are limited to only yaw and tilt with two levels of magnitude. As shown in Table~\ref{tab:object}, \name achieves strong performance across all settings, accurately reproducing the intended object motions.

\begin{table}[!h]
\centering
\caption{\textbf{Object motion evaluation on the test split of our procedural dataset.} }
\label{tab:object}
\resizebox{0.85\linewidth}{!}{
\begin{tabular}{l@{$\;\;$}c@{$\;\;$}c@{$\;\;$}c}
\toprule
 & \makecell{\textbf{Coarse}\\\textbf{Translation F1}} 
 & \makecell{\textbf{Fine}\\\textbf{Translation F1}} 
 & \makecell{\textbf{Fine}\\\textbf{Rotation F1}} \\
\midrule
Ours & \textbf{0.9983} & \textbf{0.9293} & \textbf{0.975} \\
\bottomrule
\end{tabular}
}
\end{table}

\begin{figure*}[!t]
    \centering
    \includegraphics[width=.98\textwidth]{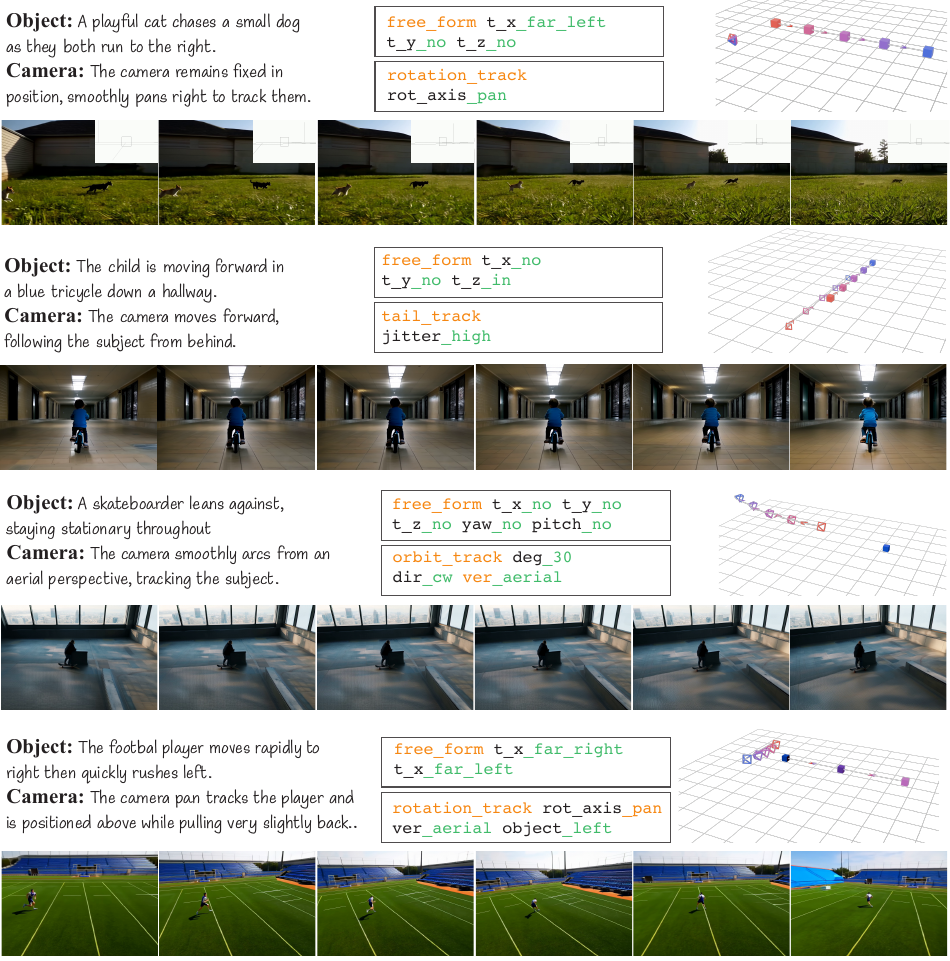}
    \caption{\textbf{Qualitative results of \name.} For each example, we show  user-provided text prompts describing object + camera motion,  corresponding DSL-based motion programs synthesized by our LLM, the resulting 3D object and camera trajectories, and generated video frames. For the first example, we also show the control video given to VACE as an inset. These showcase that \name produces coherent and physically consistent camera–object interactions across diverse  scenarios. See supplemental webpage for videos and  comparisons. }
    \label{fig:results_gallery}
\end{figure*}

\subsection{Video Evaluation}
We qualitatively evaluate \name using a diverse set of textual descriptions and the corresponding generated videos. To construct the qualitative test set, we select $\text{14}$ textual descriptions from CameraBench~\cite{camerabench}, focusing on samples involving object motion, camera motion, or both. For each case, we augment the original camera-motion captions with object-motion descriptions automatically generated by a vision–language model (VLM)~\cite{wang2025internvl3_5} prompted to describe the dynamics of the foreground objects. %To further enhance diversity, we additionally curate $\text{xxx}$ new text samples following a similar camera and object description format covering a broad range of cinematographic scenarios

\paragraph{Baselines} We generate videos for each test sample using four different settings: (i) a text-only baseline using the Wan~2.1~\cite{wan2025} video diffusion model conditioned solely on textual prompts consisting of combined object and camera motion descriptions; (ii) an in-context LLM baseline (GPT-DSL), where a state-of-the-art MLLM, GPT 5 (thinking)~\cite{gpt5} is provided with the complete specification of our DSL and prompted to generate camera-motion programs, which are then converted into trajectories and used to guide video synthesis; (iii) a similar in-context LLM baseline (GPT-Traj) where the same MLLM is given an example motion trajectory with a detailed description and is asked to synthesize motion trajectories directly; and (iv) our full model. Note that for the MLLM baselines we use the object motions generated by our method and prompt the MLLM only to generate the camera motion.

%\paragraph{Baselines}
%\begin{itemize}
%    \item provide the text prompt (structure prompts?) directly to Wan (VACE w/o control) to get a video
%    \item provide the text prompt and our camera trajectories to get a video?
%    \item ours
%\end{itemize}
%
%\paragraph{Metrics} We evaluate motion consistency and smoothness using object detection and tracking metrics following the protocol of VideoDirectorGPT~\cite{Lin2023VideoDirectorGPT}. This includes analyzing object trajectories for spatial continuity, stability, and coherence with the described actions.
%
\begin{figure}[h!]
    \centering
    \includegraphics[width=\columnwidth]{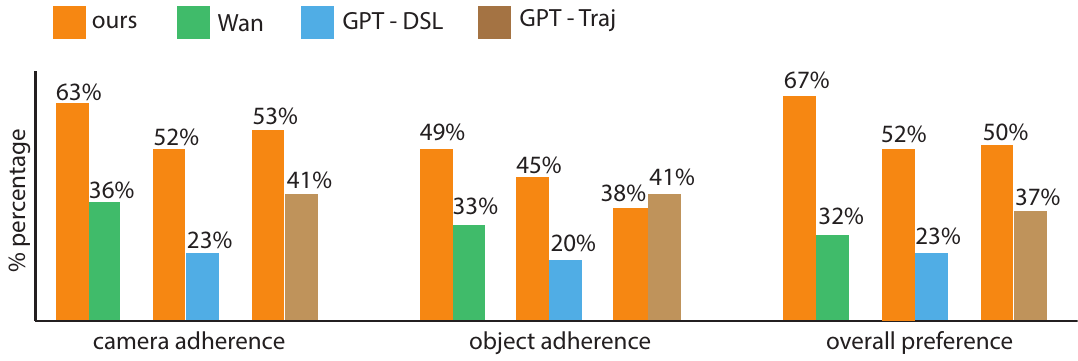}
    \caption{\textbf{User study results.} We compare \name against baseline methods in a qualitative user study. Participants consistently preferred our results in terms of camera adherence, object adherence, and overall video quality, demonstrating stronger alignment between textual descriptions and generated motion.}
    \label{fig:user_study}
\end{figure}

\paragraph{Qualitative results}
We show representative results from \name in Fig.~\ref{fig:results_gallery}, illustrating the full generation pipeline, from natural-language descriptions to DSL-based motion programs, 3D object and camera trajectories, and final rendered videos. We refer to the supplementary material for more results including the comparisons with the baselines. The examples demonstrate that \name effectively translates textual intent into coherent, physically consistent, and cinematographically meaningful motion. %As shown in Fig.~\ref{fig:results_gallery}, 
The predicted symbolic programs produce smooth trajectories that capture diverse cinematic behaviors such as following, orbiting, and free-form movement. The generated videos exhibit realistic camera–object coordination and temporal continuity. These qualitative findings complement the quantitative evaluations given earlier, confirming that symbolic motion reasoning enables stable, interpretable, and visually coherent video generation across varied scene descriptions.

\begin{table}[!h]
\centering
\caption{\textbf{DSL vs trajectory ablation on the procedural dataset.}}
\label{tab:ablation}
\resizebox{\linewidth}{!}{
\begin{tabular}{l@{$\;\;$}c@{$\;\;$}c@{$\;\;$}c@{$\;\;$}c@{$\;\;$}c}
\toprule
Our Method & \makecell{\textbf{Coarse}\\\textbf{Trans. F1}\\\textbf{(27 classes)}} 
 & \makecell{\textbf{Fine}\\\textbf{Trans. F1}\\\textbf{(343 classes)}} 
 & \makecell{\textbf{Coarse}\\\textbf{Rot. F1}\\\textbf{(27 classes)}} 
 & \makecell{\textbf{Fine}\\\textbf{Rot. F1}\\\textbf{(1728 classes)}} 
 & \makecell{\textbf{Rot. MAE}\\\textbf{per axis}\\\textbf{(degrees)}} \\
\midrule
w/ DSL & \textbf{0.996} & \textbf{0.966} & \textbf{0.950} & \textbf{0.753} & \textbf{3.507} \\
w/ traj & 0.847 & 0.781 & 0.730 & 0.626 & 8.333 \\
%w/ combined & \textul{0.961} & \textul{0.946} & \textul{0.895} & \textul{0.703} & \textul{7.272} \\
\bottomrule
\end{tabular}
}
\end{table}
\paragraph{User Study} We conduct a user study to assess perceptual alignment and cinematic quality. We ask \users participants to rate each generated video according to three criteria: (i)~alignment between camera motion and text description; (ii)~alignment between object motion and text description; and (iii)~overall plausibility of the generated video. For each example and question, we ask the users to pick one of ours, baseline result, or both if they are really indecisive. We provide the results in Fig.~\ref{fig:user_study} and refer to the supplementary for details. Our method is consistently preferred over the baselines demonstrating its effectiveness. The GPT baseline that is prompted to generate camera motion programs in the DSL format is a strong baseline (ours is preferred $52\%$ while both options are preferred $25\%$ of the time) highlighting the advantage of using a symbolic motion program representation. As expected, GPT baselines use our object motions and we do not see a significant preference for object adherence. 

%\paragraph{VLMs as a Judge} We can come up with VQA benchmark to evaluate the final videos. Some options to use: CameraBench (limited to camera only), MotionSight~\cite{du2025motionsight}, closed source options like GPT or Gemini?
%
%Use a video captioner (AuroCap) to caption the generated video, then use an LLM to compare the original and the generated caption

\subsection{Ablations}
We formulate motion planning as a symbolic program synthesis task using a DSL inspired by cinematography conventions. To examine the impact of this design, we compare it against a variant that directly predicts camera trajectories in continuous space. Specifically, using our procedural dataset focused on free-form camera motions, we fine-tune an LLM to predict per-frame trajectory parameters, where rotation and translation are discretized into bins similar to GenDoP formulation. We then evaluate the original \name model with DSL against this direct regression variant with trajectories and a combination of those on a held-out test split of the dataset. As shown in Table~\ref{tab:ablation}, the DSL-based approach yields higher directional accuracy and lower rotation error, showing that the LLM is more effective when reasoning over structured symbolic programs. Additional ablations in the supplementary compare our dataset with DataDoP and ET, showing improved cross-dataset performance.

%\begin{enumerate}
%    \item A single LLM vs two LLMs
%    \item Ablation of datasets 
%\end{enumerate}
%
%\begin{itemize}
%    \item multi-turn editing on the intermediate representations
%\end{itemize}

%% file: sections/05_conclusion.tex
\section{Conclusion}
We have presented \name, a language-driven interface for motion control, where an LLM acts as a cinematographer that writes symbolic programs in a motion language inspired by the CameraBench~\cite{camerabench} taxonomy. By grounding textual intent in interpretable motion primitives rather than raw coordinates, we enable controllable, expressive, and data-efficient generation.
%It effectively translates human language into the language of the motion, linking words, geometry, and visual storytelling within a unified generative framework.
%
%By introducing a motion domain-specific language (DSL), inspired by cinematography conventions, \name translates textual scene descriptions into structured motion programs that deterministically produce 3D trajectories for dynamic objects and relatively defined cameras. Our representation unifies object and camera control, offering an interpretable and flexible interface for composing complex and cinematic scenes simply using text prompts. 
Through large-scale procedurally generated text–trajectory data and benchmark evaluations, we demonstrate that our method substantially improves motion controllability and alignment with user intent, drastically expanding the scope of text-based directorial control over existing alternatives.  

\paragraph{Limitations} While effective, \name has limitations that reveal several directions for future work. Current DSL-based object/camera programs are limited in diversity and realism, lacking fine-grained orientation data, semantic object categories (e.g., car versus human), and rich multi-object interactions. Expanding toward dynamic, in-the-wild datasets would better capture the variety and idiosyncrasies of real-world motion. Another promising extension is allowing users to iteratively refine trajectories and recompose shots with the options of providing appearance specifications or locking identities over space and time. %Finally, integrating multimodal LLMs capable of understanding and generating across text, images, and video could enable personalized and fully context-aware scene synthesis, leading to models that not only plan trajectories but also reason about objects, appearance, and cinematic style. 
Ultimately, \name moves toward a paradigm of language-driven cinematography, where high-level textual intent directly governs motion design in generative video models.

%% file: sections/06_supplemental.tex
\clearpage
\setcounter{page}{1}
\setcounter{section}{0}
\maketitlesupplementary

This document provides additional details and results complementing the main paper. Section~\ref{sec:dataset-stats} summarizes the statistics of our procedural dataset and visualizes the distribution of motion types.
Section~\ref{sec:additional-eval-camera} presents extended quantitative evaluations on DataDoP and ET, including experiments with DSL-converted real data.
Section~\ref{sec:additional-results} shows further qualitative examples, long-horizon generation, and multi-object motion results.
Section~\ref{sec:user-study} describes the user study setup.
Section~\ref{sec:motion-dsl} provides a full specification of our motion domain-specific language (DSL).
Section~\ref{sec:failure-cases} discusses typical failure cases of the video generator and future opportunities for improving trajectory conditioning.

\section{Procedural Dataset Statistics}
\label{sec:dataset-stats}
Our procedural dataset consists of 200+K samples for camera (100K with free-form and 100K with relative motion) and 100K samples for object motion. Each object motion trajectory consists of 4 segments,
 where one motion primitive is assigned per segment,  resulting in 400K segments in total. Each sample includes a DSL program, natural-language camera caption, and the corresponding 3D trajectory. 

\begin{figure}[h!]
\begin{overpic}[width=\columnwidth]{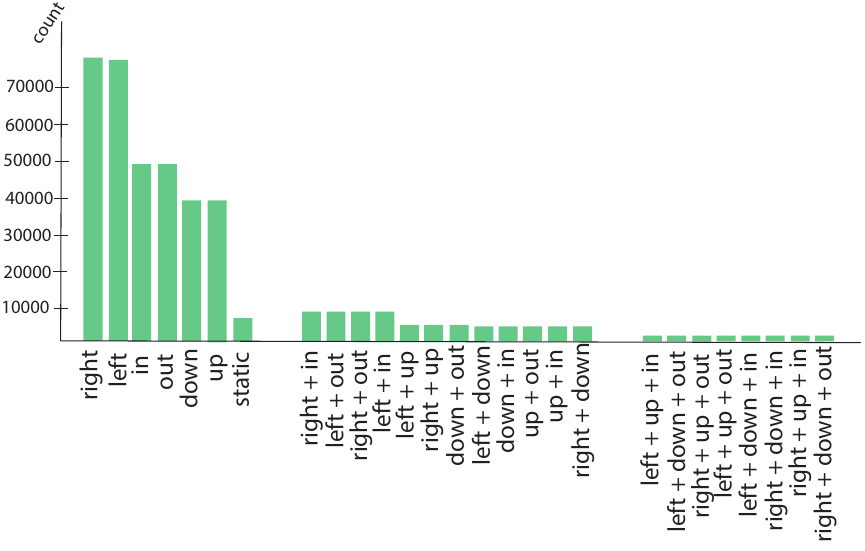} 
%\put(20,-1){(a) Free-form camera} 
%\put(77,-1){(b) Objects} 
\end{overpic}
%\vspace{1pt}
\caption{\textbf{Object motion distribution.} Frequency of single-axis and multi-axis object motions within our procedural dataset, capturing both dominant patterns and less common combinations. 
    }
\label{fig:data-object}
\end{figure}

\begin{figure}[h!]
\begin{overpic}[width=\columnwidth]{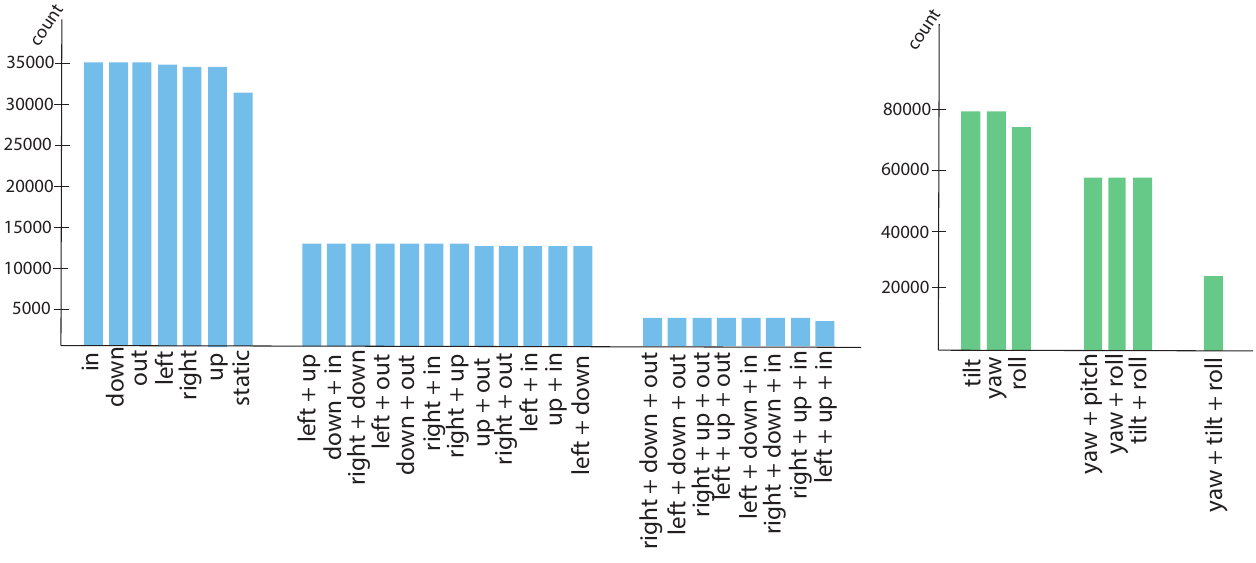} 
%\put(20,-1){(a) Free-form camera} 
%\put(77,-1){(b) Objects} 
\end{overpic}
%\vspace{1pt}
\caption{\textbf{Free-form camera motion distribution.} Frequency of single-axis and multi-axis motions for both translation (left) and rotation (right), illustrating the natural long-tailed structure of sampled camera behaviors.}
\label{fig:data-cam}
\end{figure}

Fig.~\ref{fig:data-cam} provides a detailed distribution of free-form camera motions across 27 coarse translational motion categories (3 motion types × 3 directions) together with all rotational motion combinations. Fig.~\ref{fig:data-object} shows the analogous distribution for object motions. As in real datasets~\cite{courant2024et,zhang2025gendop}, the distribution is intentionally imbalanced in that common motions single-axis motions (e.g., forward, backward) appear more frequently than multi-axis composite motions. 

\section{Additional Quantitative Evaluations for Camera Motion }
\label{sec:additional-eval-camera}
The main paper evaluates cross-dataset generalization by training \name solely on our procedural dataset and testing on DataDoP \cite{zhang2025gendop} and ET \cite{courant2024et}. Here, we analyze the inverse setting: training or finetuning \name directly on these datasets after converting their trajectories to our DSL.

\begin{figure}[!b]
\begin{overpic}[width=0.9\columnwidth]{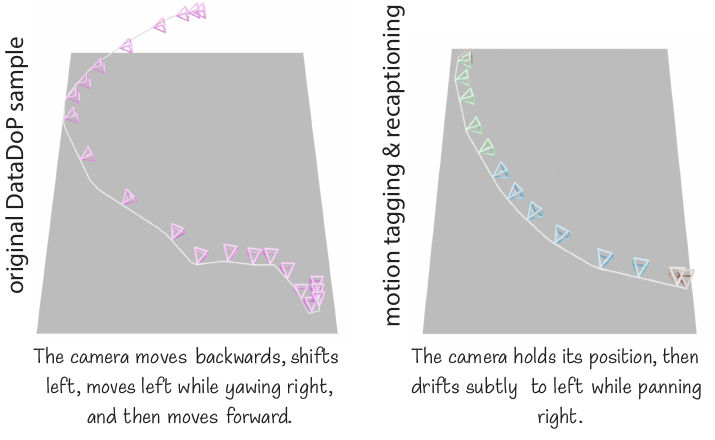} 
%\put(20,-1){(a) Free-form camera} 
%\put(77,-1){(b) Objects} 
\end{overpic}
%\vspace{1pt}
\caption{\textbf{Using real data for training.} Samples from real datasets (e.g., DataDoP) are converted to our DSL format via motion tagging and optionally re-captioned. The DSL conversion smooths the noisy trajectories extracted from real videos and improves alignment between textual descriptions and motion.}
\label{fig:data-noise}
\end{figure}

\begin{figure*}[!t]
\begin{overpic}[width=\textwidth]{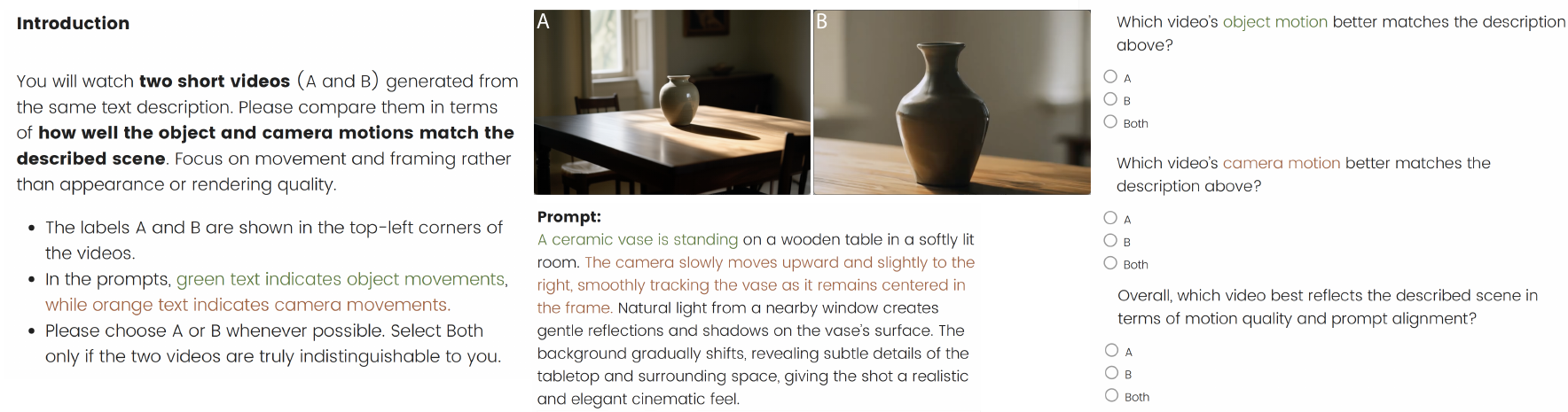} 
%\put(20,-1){(a) Free-form camera} 
%\put(77,-1){(b) Objects} 
\end{overpic}
%\vspace{1pt}
\caption{\textbf{User-study interface.} Given a prompt and a pair of videos, users compare the outputs along three criteria. 
    }
\label{fig:us_ex}
\end{figure*}

\subsection{Training on DataDoP}
DataDoP trajectories, which are extracted from real videos, are inherently noisy (see Fig.~\ref{fig:data-noise}). After DSL conversion, we compute similarity between original and DSL-reconstructed trajectories, hiscard samples below a threshold, and remove static or near-static cases. This yields 10K usable samples.
We evaluate four configurations: (i) DSL trajectories + original captions, (ii) DSL + regenerated captions, (iii)
Finetuning (\emph{ft}), and (iv) Training from scratch (\emph{tr}).

Table \ref{tab:datadop_sup} shows that while additional training data yields small gains, the improvements remain marginal due to the limited data volume. Importantly, the results confirm that our procedural dataset is already diverse and enables strong cross-dataset generalization.

%First 5 rows (numbers, scores) of table 5 is retrieved from original GenDoP paper. Other rows we experimented using official repository. However, F1-score for GenDoP was calculated on CLaTr reconstructed trajectories which is a bit odd, rather than the real and predicted traejctories directly. Former approach is getting limited by capabilities of CLaTr model, which is trained only on DataDoP. Instead, following the original approach of ET we revised this F1 score and observed a significant improvance as now it is indeed checking the direction and rotations by classes directly.

In Table 5, the results for the first five baselines are directly sourced from the original GenDoP publication, while the subsequent rows represent our experimental results obtained using the official implementation. A critical revision in our evaluation involves the F1-score computation\footnote{We have verified the necessity of this revision with the authors of GenDoP.
}. The original GenDoP framework evaluates F1-scores on trajectories reconstructed by CLaTr. However, this introduces a dependency on CLaTr's reconstruction quality, which serves as a bottleneck given its limited training on DataDoP. Aligning with the rationale established in ET, we consider direct evaluation to be more indicative of the model's true generative performance. Consequently, we revised the metric to evaluate the direction and rotation of the predicted trajectories directly. This refinement eliminates the reconstruction bias and yields a significant improvement in the observed F1-scores.

%In Table~\ref{tab:datadop_sup}, we evaluate the performance of \name when trained on the DataDoP dataset. For this purpose, we first convert the camera trajectories provided in the original dataset to our DSL format by applying motion tagging. We observe that (see Fig.~\ref{fig:data-noise}) the original data samples are noisy given that they have been recovered from real videos. Once converted to motion tags and DSL format, the trajectories are smoothed out. We compute a similarity score between the original trajectories and trajectories recovered after motion tagging and discard samples where the similarity is below a threshold. We further discard mostly static trajectories as they do not introduce much diversity. After this filtering, we use $10k$ samples from DataDoP for training. We experiment with using the additional data samples in two formats. First, we use the DSL formatted trajectories with the original captions (denoted as \emph{org cap}). Second, we also generate new captions based on the DSL programs similar to our procedural dataset. Finally, we test both finetuning \name on this additional data (denoted as \emph{ft}) and training from scratch (denotes as \emph{tr}). As shown in the table, due to the filtering required, we observe that the additional data provides only minor improvements in certain metrics. We also conclude that our procedural dataset is already diverse and demonstrates strong cross-dataset generalization. 

\begin{table}[!h]
\centering
\caption{\textbf{Camera trajectory evaluation on the DataDoP dataset.} \name achieves performance comparable to DataDoP-trained baselines despite no dataset-specific training.}
\label{tab:datadop_sup}
\resizebox{\linewidth}{!}{
\begin{tabular}{rc@{$\;\;$}c@{$\;\;$}c@{$\;\;$}c@{$\;\;$}c@{$\;\;$}c}
\toprule
\textbf{Model} & \textbf{Data} & \textbf{Revised F1-Score}&\textbf{F1-Score}  & \textbf{CLaTr} & \textbf{Coverage} & \textbf{FID} \\
\midrule 
CCD & pretrained &- &0.297 & 5.29 & 0.332 & 357.822 \\
ET & pretrained & -&0.330 & 2.46 & 0.020 & 609.906 \\
Director3D & pretrained& -& 0.058 & 0.00 & 0.171 & 542.385 \\
Director3D & DataDoP & -&0.391 & 31.69 & 0.839 & 31.979 \\
GenDoP  & DataDoP &- &\textbf{0.400} & \textbf{36.18} & \textbf{0.872} & \textbf{22.714} \vspace{0.15cm}\\
GenDoP (exp) & DataDoP & 0.360 &\textul{0.383} & 35.91 &\textbf{0.853} & \textbf{48.123}\\
Ours & pretrained & \textul{0.763} & 0.380 & \textul{36.29} & 0.794 & \textul{66.86} \\
Ours & (ft) w/ DataDoP org cap & 0.613 &0.390 & 29.44 & 0.834 & 85.46 \\
Ours & (ft) w/ DataDoP & \textbf{0.776} & \textbf{0.390} & \textbf{36.52} & 0.779 & 67.24 \\
Ours & (tr) w/ DataDoP org cap & 0.616& 0.385 & 29.13 & \textul{0.835} & 91.51 \\
Ours & (tr) w/ DataDoP & \textbf{0.776} & \textbf{0.400} & 35.48 & 0.805 & 71.52 \\
\bottomrule
\end{tabular}
}
\end{table}

\subsection{Training on ET}
We perform a similar analysis with the ET dataset. After filtering and DSL conversion, we obtain $21k$ samples from the ET dataset for additional training. Since the ET dataset lacks rotational camera motion, limiting potential gains.

Table~\ref{tab:et_sup} reports results on the pure and mixed splits. Similar to our observations with  DataDoP evaluations, DSL-converted ET data yields minor improvements when finetunes, yet \name trained purely on our procedural data already achieves the best overall generalization, further demonstrating the strength of our controlled motion design.

\begin{table}[!h]
\centering
\caption{\textbf{Camera trajectory results on the ET dataset.} The ET benchmark includes a simpler \emph{pure} and a harder \emph{mixed} split. 
\name attains the highest F1-scores on both, demonstrating strong generalization across (unseen) motion complexity levels.}
\label{tab:et_sup}
\resizebox{1.0\linewidth}{!}{

\begin{tabular}{r c@{$\;\;$}c@{$\;\;$}c@{$\;\;$}c}
\toprule

\multirow{2}{*}{\textbf{Model}} & \multicolumn{2}{c@{$\;\;$}}{\textbf{Pure Split}} & \multicolumn{2}{c}{\textbf{Mixed Split}} \\
\cmidrule(lr){2-3} \cmidrule(lr){4-5} % Pure ve Mixed'in altına çizgi ekler
& \textbf{F1 Score} & \textbf{CLaTr Score} & \textbf{F1 Score} & \textbf{CLaTr Score} \\

\midrule 

CCD & 0.27 & 3.21 & 0.17 & 6.26 \\
MDM & 0.76 & 21.26 & 0.34 & 18.32 \\
ET - DirB & 0.86 & 23.10 & 0.39 & 20.78 \\
ET - DirC & 0.80 & 21.49 & 0.48 & 21.95 \\
Ours & 0.976 & \textbf{35.10} & \textul{0.769} & \textul{36.59} \\
Ours (ft w/ ET org cap) & 0.666 & 30.66 & 0.446 & 28.06 \\
Ours (ft w/ ET) & \textul{0.978} & \textul{35.02} & \textbf{0.779} & \textbf{36.95} \\
Ours (tr w/ ET) & \textbf{0.980} & \textbf{35.10} & 0.755 & 36.37 \\
Ours (tr w/ DataDoP) & 0.967 & 34.90 & 0.747 & 35.71 \\
\bottomrule
\end{tabular}
}
\end{table}

\section{Universality of DSL Output}

\begin{figure}[!h]
    \centering
    \includegraphics[width=\columnwidth]{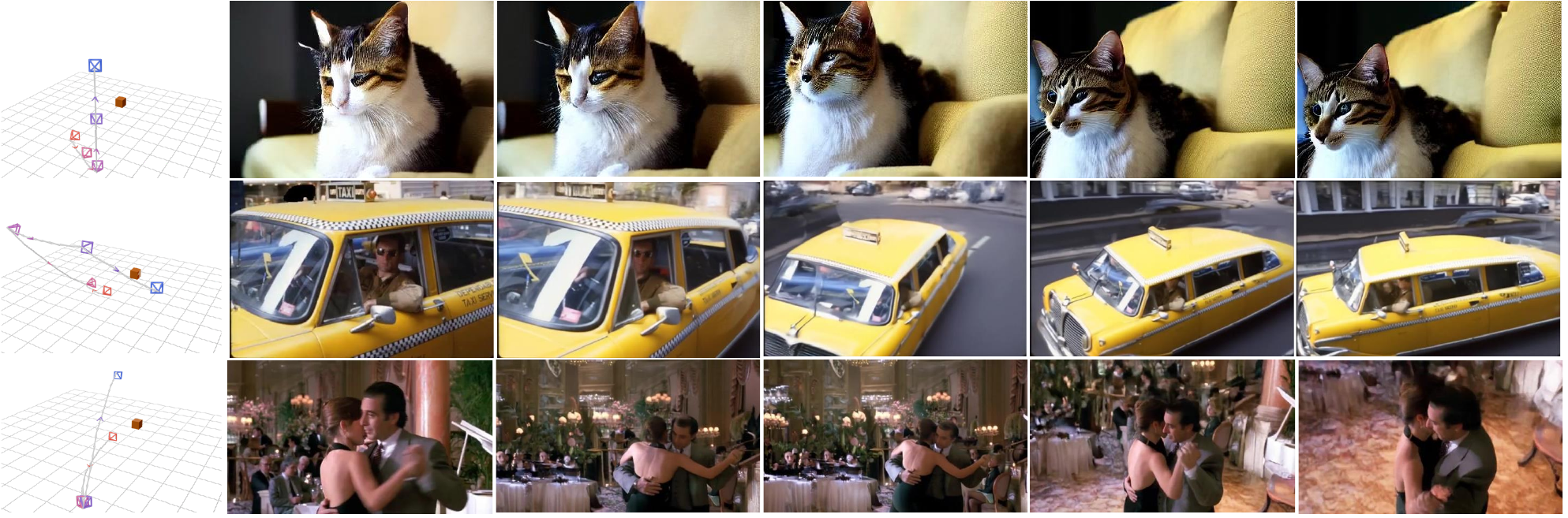}
    
    \caption{\textbf{
    DSL w/ CameraCtrl, EPiC, and ReCamMaster}. DSL applied to multiple off-the-shelf models (T2V, I2V, V2V, respectively).}%, demonstrating the model-agnostic nature of LAMP.}
    \label{fig:generators}
\end{figure}

 Since LAMP produces explicit 6-DoF camera and object trajectories, it provides a \textbf{model-agnostic interface} that does not require any modification to the video generator. We integrated the same DSL-derived motion signal with off-the-shelf video generation frameworks: \textit{text-to-video} (CameraCtrl \cite{cameractrl2}), \textit{image-to-video} (EPiC \cite{epic}), and \textit{video-to-video} (ReCamMaster \cite{recammaster}). \textit{}{As shown in Fig. \ref{fig:generators}, the identical DSL program successfully drives all three pretrained backbones}. %supporting the reviewer’s observation that it \textit{“can be integrated with off-the-shelf conditional video generation models.”}
\vspace{-0.33em}
\section{Additional Qualitative Results}
\label{sec:additional-results}
We provide extensive qualitative examples on the \hyperlink{https://cyberiada.github.io/LAMP/}{project page}.  Beyond reproducing results in the main paper, we report:

\begin{itemize}

\item \textbf{Multi-object motion:} \name can generate trajectories for multiple objects independently, followed by camera motion relative to one of them. After manually aligning initial object positions, VACE can synthesize videos conditioned on the resulting 3D paths. Automating relational multi-object trajectory generation is left for future work. Automating this step would allow LLMs to define complex temporal and spatial scene layouts and interactions directly. Since an MLLM call (Qwen2.5-VL-7B) typically takes less than 5 seconds and trajectory rendering requires only 2–3 seconds, users can validate motion in near real-time. This allows for rapid iteration before committing to the final video synthesis, which takes around 6 minutes for 81 frames with VACE-1.3B (on NVIDIA A40 48GB).

\item \textbf{Long-horizon motion generation:} We prompt \name sequentially across temporal segments. DSL-generated trajectories are concatenated by initializing each segment with the previous endpoint. VACE, which is limited to short clips, is run iteratively using the last frame as conditioning. We expect that the visual quality of the results will improve as the video models become capable of generating longer video sequences. While we anticipate that future video models will achieve higher visual fidelity for longer sequences, explicit trajectory control will remain essential. Such control is critical for orchestrating complex, multi-stage movements and ensuring precise spatiotemporal planning that simple text prompts cannot guarantee.

\end{itemize}

% \vspace{-0.5\baselineskip}

\section{User Study Details}
\label{sec:user-study}
As illustrated in Fig.~\ref{fig:us_ex}, each question in the user study presents a text prompt and two generated videos (ours vs. a baseline), and three evaluation questions. Video order is randomized per trial, and users may choose either video or select both when undecided.

\section{Limitations}
\label{sec:failure-cases}

\paragraph{Failure Rate}
We evaluated the validity of the DSL tags generated by LAMP to assess their structural and semantic integrity. The results demonstrate that LAMP is empirically robust: across four datasets (10k captions each), it shows a
\textbf{0.11\% failure rate}, with only 3 motion type confusions and 8 invalid tags per 10k samples. This highlights the stability of the DSL and deterministic grounding.

\paragraph{Video Adherence to Motion Conditioning Controls}
In our work, we use VACE~\cite{vace} as a pre-trained video generator and convert the motion trajectories to intermediate control videos. While effective, this conditioning mechanism sometimes falls short in adhering to the conditioning signals as shown in the supplementary video. Finetuning a video generator to be conditioned on directly the 3D motion trajectories can mitigate this limitation in the future.

\section{Motion Domain Specific Language}
\label{sec:motion-dsl}
In Table~\ref{tab:free-behavior}-\ref{tab:rot-behavior}, we provide the details of the motion domain specific language we propose. 
Our DSL consists of four motion primitives, each with a corresponding list of motion modifiers that parameterize that motion type.
In particular, for each motion primitive supported, we provide a list of the motion modifiers along with their description, syntax, the possible set of values, and the default value.

%\section{DSL Summary}
%We propose motion domain specific language that enables to cast motion generation as a motion program synthesis task. 
%Our DSL consists of four different motion primitives, where each primitive has its corresponding list of motion modifiers that parameterize that motion type. %In the following tables, for each motion primitive, we provide a list of motion modifiers, their descriptions, syntax, the set of values it can be set to, and its default value.

%\pagebreak

\input{sections/07_dsl_specifications}

%%%%%%%%%%%%%%%

%% file: sections/07_dsl_specifications.tex
%Our proposed domain-specific language for motion program synthesis consists of four motion primitive types. In the following, we include a detailed list of motion modifiers associated with each primitive, their description, syntax, and the possible values they can be assigned to.

\begin{table*}[!t]
\centering
\caption{Free form behaviour (\texttt{free\_form}) and its modifiers}
\renewcommand{\arraystretch}{1.2}
\setlength{\tabcolsep}{6pt}
\begin{tabular}{lp{4cm}p{2cm}p{5cm}p{1cm}}
\toprule
\textbf{Modifier} & \textbf{Description} & \textbf{Key} & \textbf{Possible Values} & \textbf{Default} \\
\midrule

  Lateral & Lateral translation & \texttt{t\_x} & \{\texttt{far\_left}, \texttt{left}, \texttt{near\_left}, \texttt{no}, \texttt{near\_right}, \texttt{right}, \texttt{far\_right}\} & \texttt{no} \\
  Vertical & Vertical translation & \texttt{t\_y} & \{\texttt{far\_down}, \texttt{down}, \texttt{near\_down}, \texttt{no}, \texttt{near\_up}, \texttt{up}, \texttt{far\_up}\} & \texttt{no} \\
  Depth & Depth translation & \texttt{t\_z} & \{\texttt{far\_in}, \texttt{in}, \texttt{near\_in}, \texttt{no}, \texttt{near\_out}, \texttt{out}, \texttt{far\_out}\} & \texttt{no} \\
  Yaw & Yaw in degrees & \texttt{yaw} & \texttt{ \{-180, -170, -160, ..., -100, -90, -85, -80, ..., -10, -5, 0, 5, 10, ..., 80, 85, 90, 100, ..., 160, 170, 180\} } & \texttt{0} \\
  Pitch & Pitch in degrees & \texttt{pitch} & \texttt{ \{-180, -170, -160, ..., -100, -90, -85, -80, ..., -10, -5, 0, 5, 10, ..., 80, 85, 90, 100, ..., 160, 170, 180\} } & \texttt{0} \\
  Roll & Roll in degrees & \texttt{roll} & \texttt{ \{-180, -170, -160, ..., -100, -90, -85, -80, ..., -10, -5, 0, 5, 10, ..., 80, 85, 90, 100, ..., 160, 170, 180\} } & \texttt{0} \\

\bottomrule
\end{tabular}
\label{tab:free-behavior}
\end{table*}

\begin{table*}[!t]
\centering
\caption{Orbit track behaviour (\texttt{orbit\_track}) and its modifiers}
\renewcommand{\arraystretch}{1.2}
\setlength{\tabcolsep}{6pt}
\begin{tabular}{lp{4cm}p{2cm}p{5cm}p{1cm}}
\toprule
\textbf{Modifier} & \textbf{Description} & \textbf{Key} & \textbf{Possible Values} & \textbf{Default} \\
\midrule

  Dutch & Camera roll angle & \texttt{dutch} & \texttt{ \{-45, -30, -15, 0, 15, 30, 45\} } & \texttt{0} \\
  Easing & Defines the camera acceleration curve & \texttt{ease} & \{\texttt{in}, \texttt{out}, \texttt{in\_out}, \texttt{out\_in}, \texttt{linear}\} & \texttt{linear} \\
  Jitter & Small, random vibrations, simulating a handheld effect & \texttt{jitter} & \{\texttt{low}, \texttt{high}, \texttt{none}\} & \texttt{none} \\
  Vertical Angle & Camera's vertical perspective relative to the object & \texttt{ver} & \{\texttt{aerial}, \texttt{low-angle}, \texttt{none}\} & \texttt{none} \\
  Framing Offset & Offsets the object’s position within the frame & \texttt{object} & \{\texttt{left}, \texttt{right}, \texttt{none}\} & \texttt{none} \\
  Orbit Plane & The primary axis around which the camera orbits the object & \texttt{plane\_axis} & \texttt{ \{x, y, z\} } & \texttt{y} \\
  Orbit Degrees & Total angular distance & \texttt{deg} & \texttt{ \{30, 45, 60, 90, 180, 270, 360\} } & \texttt{90} \\
  Direction & Direction of rotation & \texttt{dir} & \texttt{ \{cw, ccw\} } & \texttt{cw} \\
  Spiral Dolly & Combines the orbit motion with a simultaneous camera movement towards object, creating a spiral & \texttt{spiral} & \texttt{ \{in\_0.1,  in\_0.3, in\_0.5, out\_0.1,  out\_0.3, out\_0.5, no\} } & \texttt{no} \\

\bottomrule
\end{tabular}
\label{tab:orbit-behavior}
\end{table*}

\begin{table*}[!t]
\centering
\caption{Tail track behaviour (\texttt{tail\_track}) and its modifiers}
\renewcommand{\arraystretch}{1.2}
\setlength{\tabcolsep}{6pt}
\begin{tabular}{lp{4cm}p{2cm}p{5cm}p{1cm}}
\toprule
\textbf{Modifier} & \textbf{Description} & \textbf{Key} & \textbf{Possible Values} & \textbf{Default} \\
\midrule

  Dutch & Camera roll angle & \texttt{dutch} & \texttt{ \{-45, -30, -15, 0, 15, 30, 45\} } & \texttt{0} \\
  Easing & Defines the camera acceleration curve & \texttt{ease} & \{\texttt{in}, \texttt{out}, \texttt{in\_out}, \texttt{out\_in}, \texttt{linear}\} & \texttt{linear} \\
  Jitter & Small, random vibrations, simulating a handheld effect & \texttt{jitter} & \{\texttt{low}, \texttt{high}, \texttt{none}\} & \texttt{none} \\
  Vertical Angle & Camera's vertical perspective relative to the object & \texttt{ver} & \{\texttt{aerial}, \texttt{low-angle}, \texttt{none}\} & \texttt{none} \\
  Framing Offset & Offsets the object’s position within the frame & \texttt{object} & \{\texttt{left}, \texttt{right}, \texttt{none}\} & \texttt{none} \\
  Follow style & Responsiveness of the camera (strict vs delayed) & \texttt{follow\_style} & \texttt{ \{hard, soft, lazy\} } & \texttt{hard} \\
  Follow axis & Axis or axes the camera uses to follow the object & \texttt{follow\_axis} & \texttt{ \{x, y, z, full\} } & \texttt{full} \\
  Amplitude & Scales the camera’s travel distance relative to the object & \texttt{amp} & \texttt{ \{x\_0.5, x\_0.8, x\_1.2, x\_1.5, y\_0.5, y\_0.8, y\_1.2, y\_1.5, z\_0.5, z\_0.8, z\_1.2, z\_1.5, all\_0.5, all\_0.8, all\_1.2, all\_1.5, no\} } & \texttt{no} \\
  Static Dolly & Moves the camera toward or away from the object & \texttt{dolly} & \texttt{ \{in\_0.1,  in\_0.3, in\_0.5, out\_0.1,  out\_0.3,  out\_0.5, no\} } & \texttt{no} \\
  Mirror & Produces symmetrical camera motion & \texttt{mirror\_axis} & \texttt{ \{x, y, no\} } & \texttt{no} \\
  Look at & Disables or enforces orientation towards the objects & \texttt{dont\_look} & \texttt{ \{dont\_look, none\} } & \texttt{none} \\
  Lead & Positions the camera ahead of the object’s motion direction,& \texttt{lead} & \texttt{\{lead, none}\} & \texttt{none} \\

\bottomrule
\end{tabular}
\label{tab:tail-behavior}
\end{table*}

\begin{table*}[!t]
\centering
\caption{Rotation track behaviour (\texttt{rotation\_track}) and its modifiers}
\renewcommand{\arraystretch}{1.2}
\setlength{\tabcolsep}{6pt}
\begin{tabular}{lp{4cm}p{2cm}p{5cm}p{1cm}}
\toprule
\textbf{Modifier} & \textbf{Description} & \textbf{Key} & \textbf{Possible Values} & \textbf{Default} \\
\midrule

  Dutch & Camera roll angle & \texttt{dutch} & \texttt{ \{-45, -30, -15, 0, 15, 30, 45\} } & \texttt{0} \\
  Easing & Defines the camera acceleration curve & \texttt{ease} & \{\texttt{in}, \texttt{out}, \texttt{in\_out}, \texttt{out\_in}, \texttt{linear}\} & \texttt{linear} \\
  Jitter & Adds small, random vibrations, simulating a handheld effect & \texttt{jitter} & \{\texttt{low}, \texttt{high}, \texttt{none}\} & \texttt{none} \\
  Vertical Angle & Camera's vertical perspective relative to the object & \texttt{ver} & \{\texttt{aerial}, \texttt{low-angle}, \texttt{none}\} & \texttt{none} \\
  Framing Offset & Offsets the object’s position within the frame & \texttt{object} & \{\texttt{left}, \texttt{right}, \texttt{none}\} & \texttt{none} \\
  Rotation axis & Determines the rotation axis or axes & \texttt{rot\_axis} & \texttt{ \{pan, tilt, full\} } & \texttt{full} \\
  Local Dolly & Controls the distance of the camera to the object during tracking & \texttt{push} & \texttt{ \{in\_0.1,  in\_0.3, in\_0.5, out\_0.1,  out\_0.3,  out\_0.5, no\} } & \texttt{no} \\
  Local Offset & Shifts the camera’s look-at point relative to the target & \texttt{local\_offset} & \texttt{ \{x\_-0.3, x\_-0.1, x\_0.1, x\_0.3, y\_-0.3, y\_-0.1, y\_0.1, y\_0.3, no\} } & \texttt{no} \\
  World moves 1 & If enabled, the camera rotates while compensating for world-space motion to maintain focus on the target. In this mode, local modifiers are not used. &  & \texttt{\{truck\_right\_\{amount\}, truck\_left\_\{amount\}, pedestal\_up\_\{amount\}, pedestal\_down\_\{amount\}, goes\_in\_\{amount\}, goes\_out\_\{amount\} } & \texttt{none} \\
  World moves 2 & Similar to \emph{World moves 1} but defines the world-space motion in the second half of the shot &  & \texttt{\{truck\_right\_\{amount\}, truck\_left\_\{amount\}, pedestal\_up\_\{amount\}, pedestal\_down\_\{amount\}, goes\_in\_\{amount\}, goes\_out\_\{amount\} } & \texttt{none} \\

\bottomrule
\end{tabular}
\label{tab:rot-behavior}
\end{table*}